\documentclass[runningheads]{llncs}

\usepackage{eccv}

\usepackage{eccvabbrv}

\usepackage{graphicx}
\usepackage{booktabs}
\usepackage{subcaption}
\usepackage[accsupp]{axessibility}  

\usepackage{hyperref}

\usepackage{orcidlink}

\begin{document}

\title{Multimodal Fusion Strategies for Mapping Biophysical Landscape Features}

\author{Lucia Gordon\inst{1,2}\orcidlink{0000-0003-3219-6960} \and
Nico Lang\inst{2}\orcidlink{0000-0001-8434-027X} \and
Catherine Ressijac\inst{1} \and Andrew Davies\inst{1}\orcidlink{0000-0002-0003-1435}}

\authorrunning{L.~Gordon et al.}

\institute{Harvard University, USA \and
University of Copenhagen, Denmark\\
\email{\href{mailto:luciagordon@g.harvard.edu}{luciagordon@g.harvard.edu}}}

\maketitle

\begin{abstract}
  Multimodal aerial data are used to monitor natural systems, and machine learning can significantly accelerate the classification of landscape features within such imagery to benefit ecology and conservation. It remains under-explored, however, how these multiple modalities ought to be fused in a deep learning model. As a step towards filling this gap, we study three strategies (\textsc{Early fusion}, \textsc{Late fusion}, and \textsc{Mixture of Experts}) for fusing thermal, RGB, and LiDAR imagery using a dataset of spatially-aligned orthomosaics in these three modalities. In particular, we aim to map three ecologically-relevant biophysical landscape features in African savanna ecosystems: rhino middens, termite mounds, and water. The three fusion strategies differ in whether the modalities are fused early or late, and if late, whether the model learns fixed weights per modality for each class or generates weights for each class adaptively, based on the input. Overall, the three methods have similar macro-averaged performance with \textsc{Late fusion} achieving an AUC of 0.698, but their per-class performance varies strongly, with \textsc{Early fusion} achieving the best recall for middens and water and \textsc{Mixture of Experts} achieving the best recall for mounds.\footnote{The code is available at \href{https://github.com/lgordon99/fusion-strategies-eccv}{https://github.com/lgordon99/fusion-strategies-eccv}.}
\end{abstract}

\section{Introduction}
\label{sec:introduction}
In response to widespread land degradation and biodiversity loss, in its Fifteenth Sustainable Development Goal ``Life on Land,'' the United Nations calls for protecting, restoring, and promoting the sustainable use and management of terrestrial ecosystems~\cite{UN}. Successfully conserving habitats and species necessitates effective ecosystem and biodiversity monitoring~\cite{parrish}. Towards this goal, ecologists and protected area managers map biophysical landscape features, often through a combination of remote sensing and ground surveys.

In African savanna ecosystems, three examples of such features of interest are rhino middens, termite mounds, and water.
Rhino middens are communal defecation sites used by rhinos for territorial marking and social communication~\cite{owen1973behavioural}. Termite mounds, composed of soil, termite saliva, and dung, are islands of enhanced nutrient and moisture content in the landscape, facilitating the growth of vegetation, which in turn attracts herbivores~\cite{loveridge}. Water, which is here defined as a river, stream, or watering hole, is a key resource on which many animal species are dependent. These landscape features are often distributed throughout vast, inaccessible areas, making manual annotation of remotely sensed imagery tedious and a complete ground survey beyond available time and resources. This challenge motivates the use of deep learning (DL) to automate the detection of biophysical landscape features in UAV imagery.

\begin{figure}[tb]
    \centering
    \includegraphics[width=0.85\linewidth]{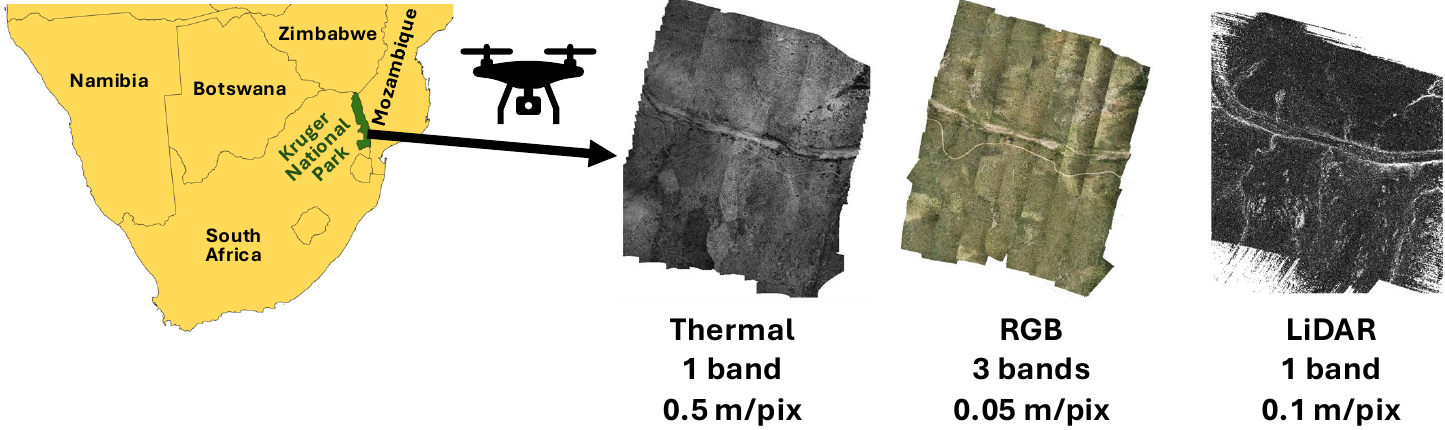}
    \caption{
    Multi-band, multi-resolution aerial imagery collected for a site in Kruger National Park, South Africa.}
    \label{fig:imagery_dataset}
\end{figure}

We develop DL models to map these features in aerial imagery taken of a 284-hectare site in Kruger National Park, South Africa in January 2020~\cite{gordon}. The UAV was equipped with a thermal camera, an RGB camera, and a LiDAR (light detection and ranging) scanner, which simultaneously collected thermal, RGB, and elevation data at 0.5-m, 0.05-m, and 0.1-m resolution, respectively (see \Cref{fig:imagery_dataset}). Using this multimodal data, we study the question of: \textbf{How to most effectively fuse multi-resolution thermal, RGB, and LiDAR data in a DL model for multi-class image classification using limited, imbalanced training data?} In particular, is it better to perform fusion earlier or later in the feature extraction process? And how should the modalities be weighted when fused later?

\section{Related Work}
This work is an extension of Gordon \etal~\cite{gordon}, which develops an active learning methodology for the detection of rhino middens in this same imagery. While the purpose of that work was to demonstrate the success of a domain-inspired active learning technique, this work explores different fusion methods and architectures and also generalizes to a multi-class setting by considering termite mounds and water in addition to rhino middens.
Burke \etal~\cite{burke} summarizes two general fusion strategies. In \emph{early fusion}, multiple input modalities are stacked and fed into a single neural network as multiple input channels. In contrast, \emph{late fusion} involves using a separate network for each of the modalities to extract modality-specific features. The features extracted by these multiple networks are then concatenated and fed into a prediction layer.

Yeh \etal~\cite{yeh} compares the performance of early and late fusion for predicting wealth across Africa from Landsat satellite imagery and nightlight data. In their early fusion method they stacked the Landsat and nightlight bands and passed them through a single network, while in their late fusion method they used two separate ResNet-18 models, one for the Landsat imagery and one for the nightlight data, to extract features from each of the modalities that were then merged and passed through a fully connected layer. They found that late fusion outperformed early fusion. 
Late fusion can be beneficial when different modalities are more or less useful for classifying images belonging to different classes.
For example, Gordon \etal~\cite{gordon} found that thermal imagery was most useful for detecting rhino middens due to their strong heat signature.
In this dataset, while the height of termite mounds makes them stand out in LiDAR imagery, the same is not true of water, which is at ground level.

A \emph{mixture of experts} in the context of DL~\cite{eigen} contains separate neural networks, or ``experts,'' that specialize in different parts of the input space, which here correspond to modalities. The outputs of the different networks are combined using weights specified by a \emph{gating network}, which maps each input to a distribution over the experts using a multilayer perceptron (MLP) with a final softmax layer to normalize the resultant weight values. The expert outputs weighted by these gating weights then serve as the final predictions. In contrast to late fusion that has fixed modality-specific weights for each class, using a gating network makes it possible to weight the modalities differently for different inputs. This is useful, for example, if some termite mounds are tall and others are short, meaning that LiDAR will be more useful for the former than the latter.

\section{Dataset}
\begin{figure}[tb]
    \centering
    \begin{subfigure}{0.45\linewidth}
        \centering
        \includegraphics[width=0.6\linewidth]{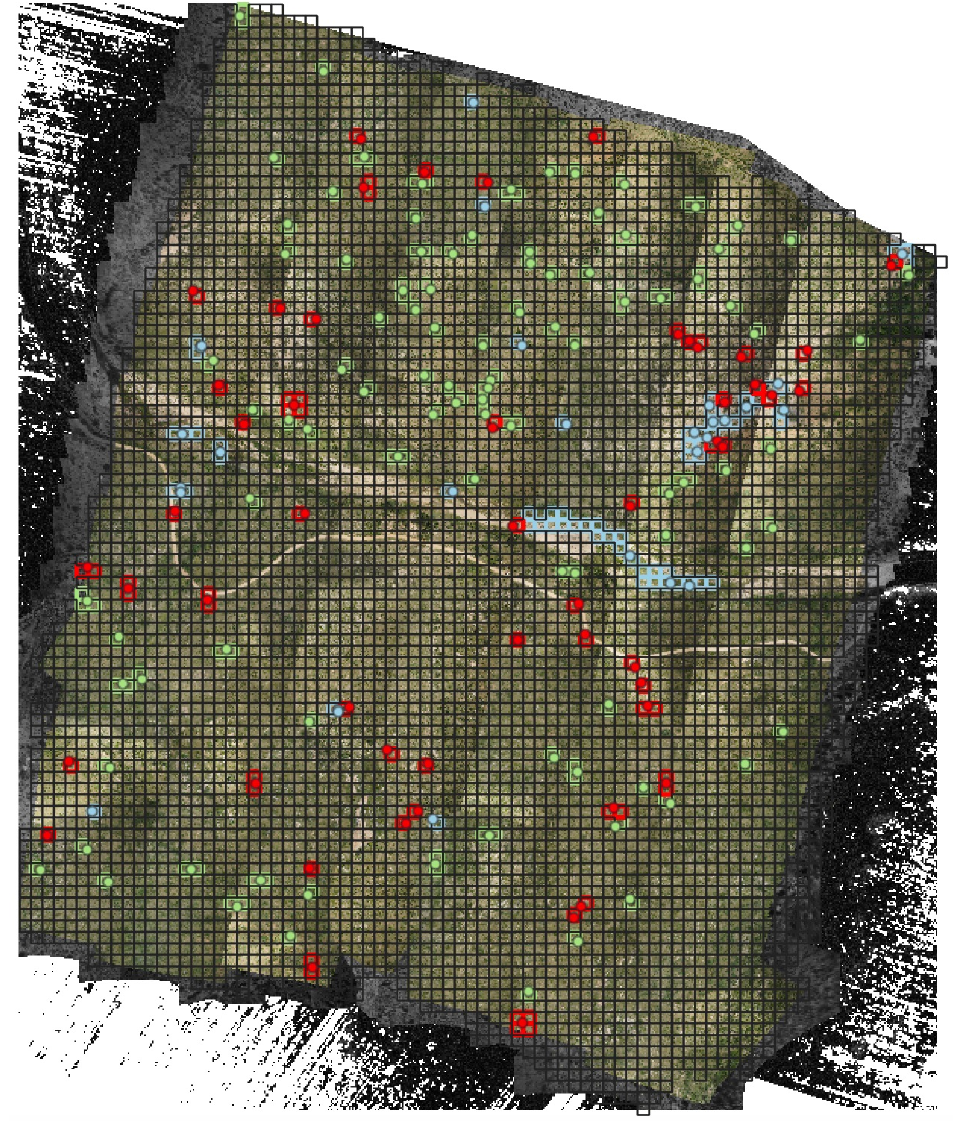}
        \caption{Thermal, RGB, and LiDAR gridded orthomosaics with landscape features: \color{red}rhino middens\color{black}, \color{green}termite mounds\color{black}, and \color{cyan}water\color{black}.}
        \label{fig:labeled_dataset}
    \end{subfigure}
    \hfill
    \begin{subfigure}{0.45\linewidth}
        \centering
        \includegraphics[width=0.65\linewidth]{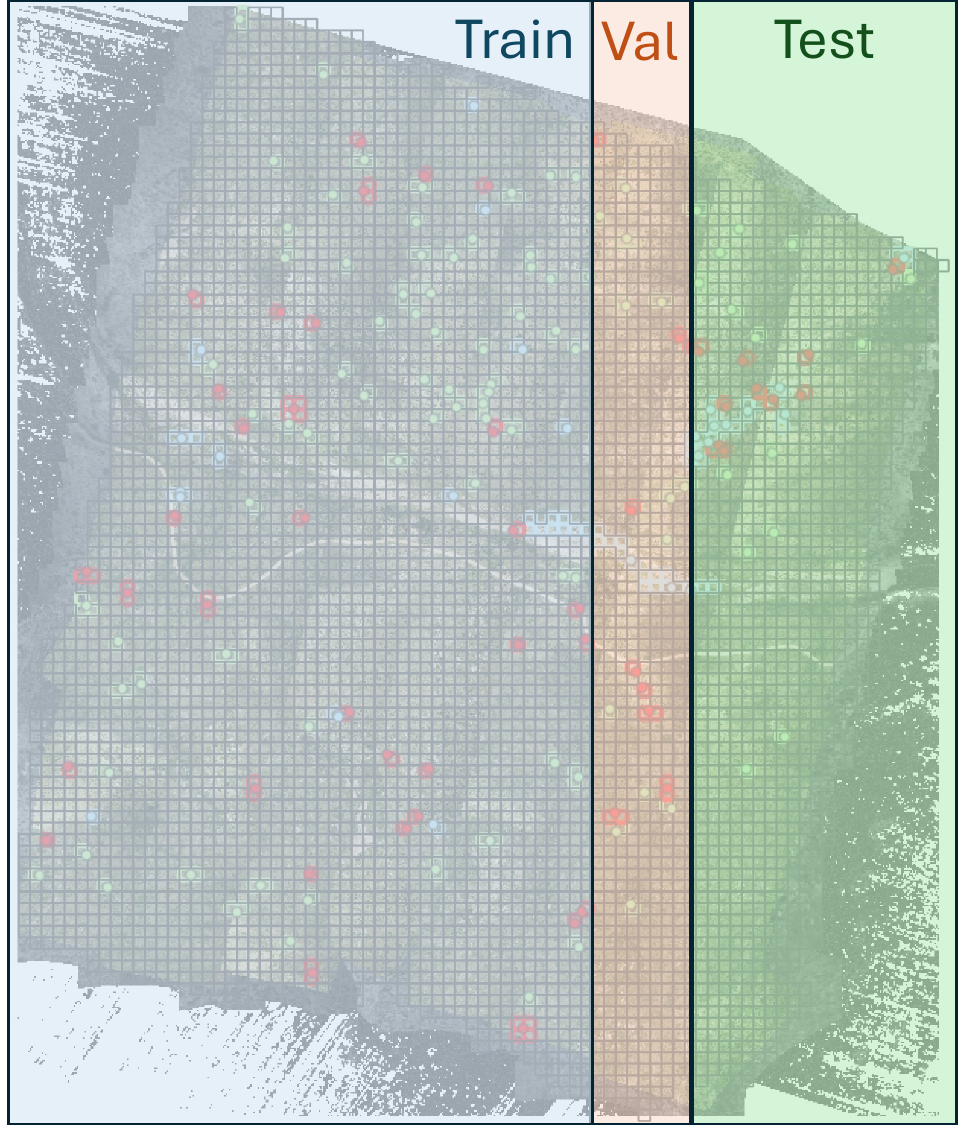}
        \caption{The site is spatially split into training, validation, and testing areas.}
        \label{fig:train_test_split}
    \end{subfigure}
    \caption{Spatial distribution of landscape features (a) and regional data split (b).}
\end{figure}

The dataset consists of spatially aligned thermal, RGB, and LiDAR orthomosaics in GeoTIFF format along with latitude-longitude coordinates for the rhino middens, termite mounds, and water bodies in the site, which we gridify into 20x20-m cells (see \Cref{fig:labeled_dataset}).
We use QGIS to assign each grid cell a label corresponding to the aforementioned landscape feature it contains, and it is labeled ``empty'' if it contains none of them.
Because the band number and resolution of the orthomosaics vary across modalities, the same cell is represented by a different-sized tile in each of the modalities: (40,40) for thermal, (3,400,400) for RGB, and (200,200) for LiDAR.
As this is a geospatial dataset, we perform a spatially-aware train-validation-test partition~\cite{rolf} (see \Cref{fig:train_test_split}), taking the first 50 columns of the grid for training, the next 9 for validation, and the remaining 22 for testing.
To address the severe class imbalance (most cells are empty), within the training set we randomly undersample grid cells in the empty class and oversample cells in the midden and water classes so that there are 88 cells for each class. The validation set contains 704 empty cells, 12 midden cells, 18 mound cells, and 17 water cells. The test set contains 1,036 empty cells, 10 midden cells, 16 mound cells, and 21 water cells. The dataset is proprietary and sensitive due to rhino poaching but could be made available upon request.

\section{Methods}
\begin{figure*}
  \centering
  \begin{subfigure}{0.4\linewidth}
    \includegraphics[width=\linewidth]{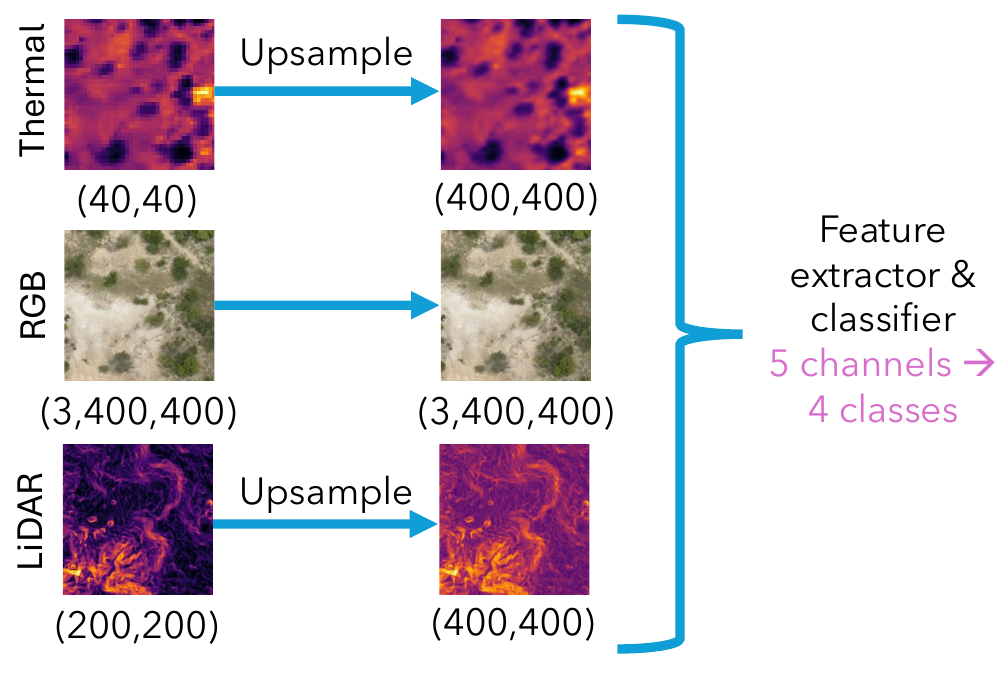}
    \caption{\textsc{Early fusion}}
    \label{fig:early_fusion}
  \end{subfigure}
  \hfill
  \begin{subfigure}{0.38\linewidth}
    \centering
    \includegraphics[width=\linewidth]{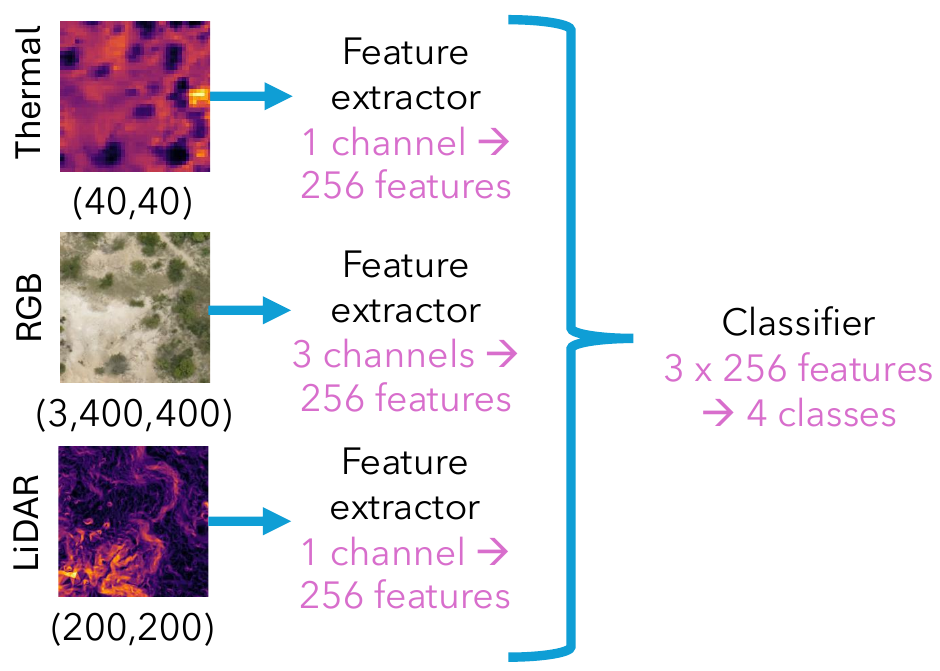}
    \caption{\textsc{Late fusion}}
    \label{fig:late_fusion}
  \end{subfigure}
  \begin{subfigure}{0.65\linewidth}
    \centering
    \includegraphics[width=\linewidth]{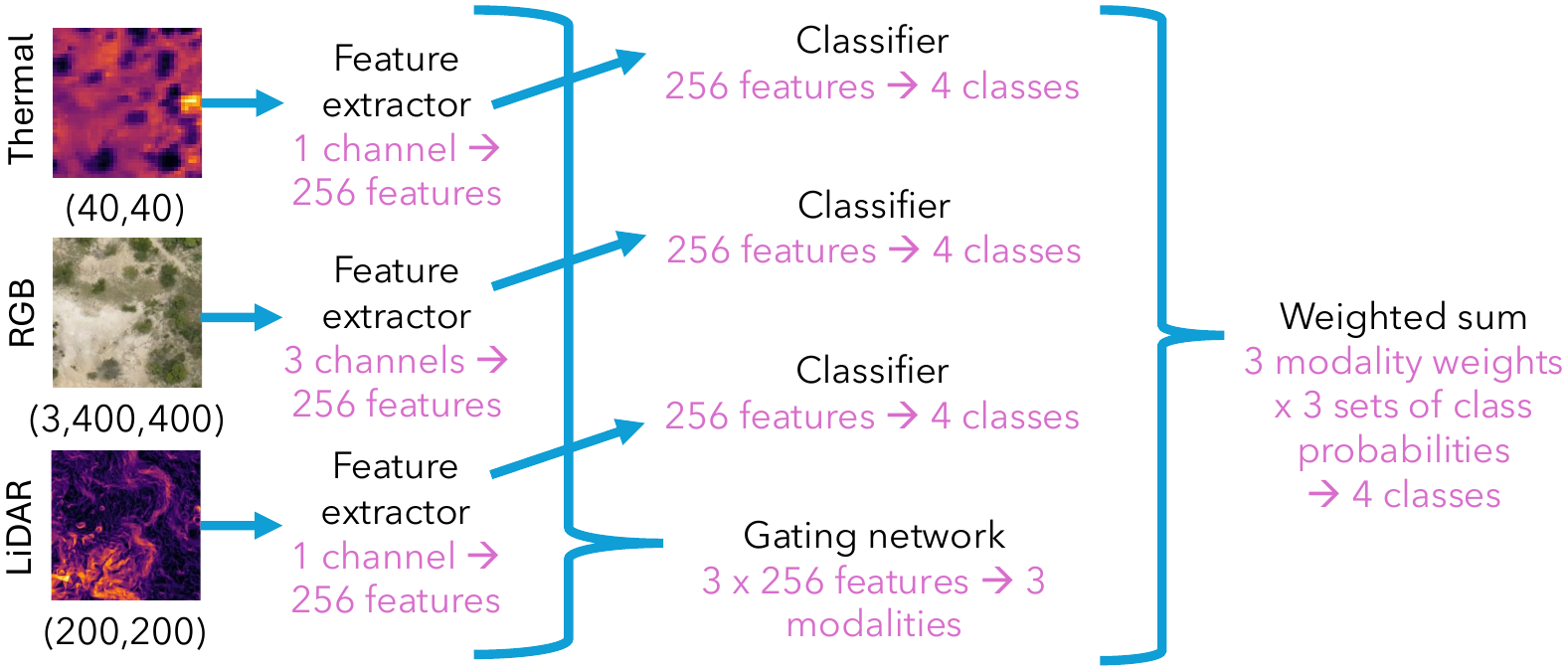}
    \caption{\textsc{Mixture of Experts}}
    \label{fig:mixture_of_experts}
  \end{subfigure}
  \caption{Diagrams for the three fusion methods studied.}
  \label{fig:fusion_methods}
\end{figure*}

To address the limited training data (88 cells per class) available, we utilize transfer learning. All our models are based on a ResNet-50~\cite{he} pretrained on ImageNet-1k~\cite{imagenet}, a dataset of ground-level RGB images. ResNet-50 takes in three-channel images and outputs a vector of class predictions. We modify the final fully connected layer in the architecture to yield a four-dimensional output to match the four classes in our dataset. We adapt the architecture further in various ways to implement three distinct fusion methods visualized in \Cref{fig:fusion_methods}.

\subsubsection{\textsc{Early fusion}.}
The \textsc{early fusion} strategy (see \Cref{fig:early_fusion}) first involves upsampling the thermal and LiDAR TIFFs using cubic resampling to match the resolution of the RGB TIFF. Each grid cell then corresponds to three tiles (thermal, RGB, LiDAR) of shapes (400,400), (3,400,400), and (400,400), respectively. We concatenate these three tiles to form a (5,400,400) array for each grid cell. Because the pre-trained ResNet-50 architecture accepts a three-channel input, we modify the first convolutional layer to accept five channels and set the weights of the non-RGB channels to the mean of the weights of the RGB channels and rescale all the weights in that first layer by 3/5, following the procedure in Yeh~\etal \cite{yeh}.

\subsubsection{\textsc{Late fusion}.}
The \textsc{Late fusion} strategy (see \Cref{fig:late_fusion}) passes the tiles of different modalities through separate feature extractors.
Since the thermal and LiDAR tiles have a single band, we modify the first convolutional layer of a ResNet-50 to accept a single-channel input.
We set the weights of that channel to the mean of the weights of the RGB channels and rescale all the weights in that first layer by 3, consistent with the procedure described for \textsc{Early fusion}.
The final fully connected layer of all three networks is modified to output 256 features.
The three sets of 256 features output by the three feature extractors are then concatenated and fed into a final fully connected layer that outputs class predictions. 

\subsubsection{\textsc{Mixture of Experts}.}
The \textsc{Mixture of Experts} (MoE, see \Cref{fig:mixture_of_experts}) method first passes the tiles of different modalities through separate feature extractors as in \textsc{Late fusion}. Next, the features extracted for each modality are passed through separate fully connected layers that output class predictions. The three sets of 256 features are also concatenated and passed through a gating network consisting of a fully connected layer outputting 256 features, a ReLU function, a fully connected layer with a three-dimensional output (for the three modalities), and a softmax function. The output of the gating network are then normalized weights on each of the modalities. 
These gating weights are used to combine the individual expert predictions in a weighted average to yield the final class predictions. 

\section{Experiments}

\subsubsection{Setup.}
We use a batch size of 64 and continue training until the AUC on the validation set has been less than the best AUC for more than 10 consecutive epochs. We normalize all the imagery using the training statistics such that each band in the training imagery has mean 0 and standard deviation 1.
We perform 50 trials for each method, where the random seed for trial $i$ is $i$.
Training for \textsc{Early fusion} and \textsc{Late fusion} utilizes cross entropy loss and \textsc{Mixture of Experts} uses negative log likelihood loss.
We use an Adam optimizer with a learning rate of 0.001 for \textsc{Early fusion} and \textsc{Late fusion}, and 0.0001 for \textsc{Mixture of Experts}.
We tuned these learning rates by selecting the best AUC on the validation set during the training period when using a learning rate of 0.1, 0.01, 0.001, 0.0001, and 0.00001.
All model parameters are fine-tuned during training. All experiments were carried out using one NVIDIA A100 SXM4 40GB graphics card. \textsc{Early fusion} required 40GB of RAM and \textsc{Late fusion} and \textsc{Mixture of Experts} required 20GB.

\subsubsection{Results.}
\begin{figure*}[tb]
  \centering
  \begin{subfigure}{0.29\linewidth}
  \centering
   \includegraphics[width=\linewidth]{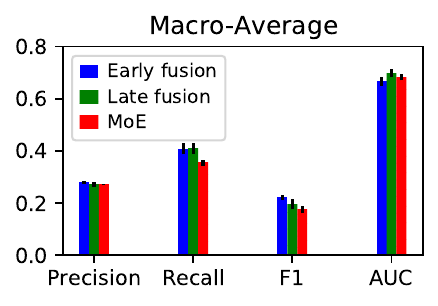}
   \caption{}
   \label{fig:overall_results}
   \end{subfigure}
  \begin{subfigure}{0.29\linewidth}
  \centering
    \includegraphics[width=\linewidth]{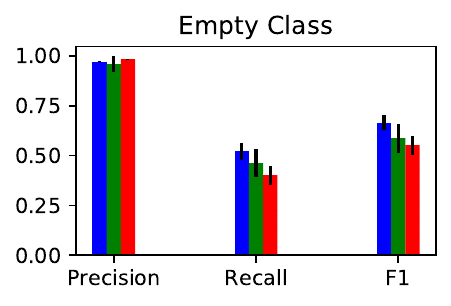}
    \caption{}
    \label{fig:empty_results}
  \end{subfigure}
  \\
  \begin{subfigure}{0.29\linewidth}
    \centering
    \includegraphics[width=\linewidth]{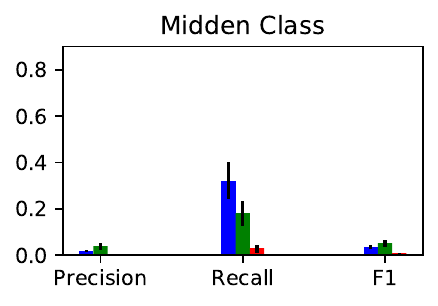}
    \caption{}
    \label{fig:midden_results}
  \end{subfigure}
  \begin{subfigure}{0.29\linewidth}
    \centering
    \includegraphics[width=\linewidth]{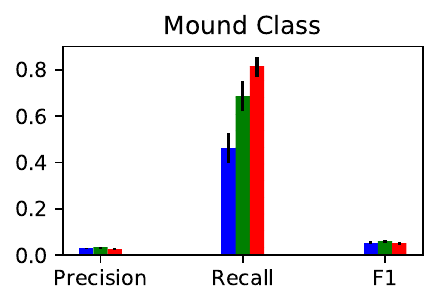}
    \caption{}
    \label{fig:mound_results}
  \end{subfigure}
  \begin{subfigure}{0.29\linewidth}
    \centering
    \includegraphics[width=\linewidth]{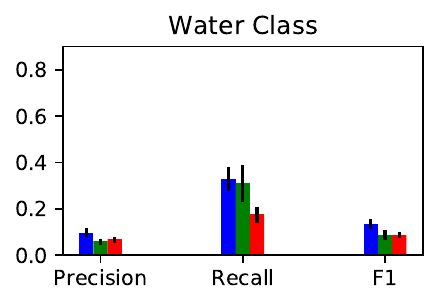}
    \caption{}
    \label{fig:water_results}
  \end{subfigure}
  \caption{Performance evaluated for three fusion methods both macro-averaged over all classes (a) and stratified by class (b--e). Mean and two standard errors are displayed.}
  \label{fig:class_results}
\end{figure*}

The overall performance of the three methods is plotted in \Cref{fig:overall_results}, which shows all three methods have very similar precision and AUC on the test set.
Comparing Figures~\ref{fig:empty_results}-\ref{fig:water_results} shows that even after balancing classes in the training set, precision was by far highest on the empty class for all methods and rather low for all landscape features, which means that all methods lead to a high number of false positives, which could be attributed to the low number of instances available for training for each class.
Recall is highest for termite mounds, with \textsc{Mixture of Experts} achieving the best performance.
We hypothesize that this is because different termite mounds can look very different in the various modalities depending on their height and vegetation, so \textsc{Mixture of Experts}, which applies input-specific modality weights, is well-suited.
We observe a large performance difference between middens and mounds, and mounds' higher recall could be due to the larger number of distinct images in the training set.
The ranking of the three methods on recall is also reversed for middens and mounds, which could be due to the approximately 3x fewer parameters in the \textsc{Early fusion} model being a better fit to the smaller amount of distinct midden instances in the training data.
\textsc{Mixture of Experts} does not introduce many more parameters than \textsc{Late fusion}, but the adaptive modality weighting might lead to overfitting for the midden and water classes. 
Though \textsc{Late fusion} does not perform best for any of the individual classes, it generalizes best across them, leading to its best AUC overall.
For all three landscape features we see that recall is much higher than precision.
For this application, recall is more important than precision since ecologists would rather have to manually discard some empty images than miss some instances of the landscape features. 
Nevertheless, future work could explore data augmentation and alternative class-balancing strategies \cite{kang} in order to improve the F1 score for all methods.

\textsc{Mixture of Experts} is the most interpretable among these fusion methods, as we can inspect the gating weights assigned to each of the modalities for test images belonging to different classes, visualized in \Cref{table:gating_weights}.
We see that for midden images, the LiDAR modality is weighted highest.
While middens show slight depressions in the LiDAR imagery due to rhinos' scraping behavior, their warmth gives them the strongest signal in thermal data \cite{gordon}, so MoE's poor midden recall could be explained by it not having learned to weight thermal highest for this class. 
For mound images, the LiDAR modality is weighted highest, even more strongly than for middens.
This is expected because mounds tend to be even taller than middens but often resemble other vegetation in thermal and RGB.
MoE achieves high recall for mounds, so its intuitive modality weighting is consistent with strong performance.
For water images, the thermal modality is weighted highest.
While water's cooler temperature gives it some signal in thermal data, we expect its blue color visible in the RGB imagery to be an even stronger signal.
Incorporating new data from other sites or using data augmentation techniques could help MoE learn to more appropriately weight the modalities for the midden and water classes.

\begin{table}[tb]
\caption{\textsc{Mixture of Experts} per-modality gating weights averaged across test images in each class and pooled across trials. Mean and two standard errors are displayed.}
\centering
\setlength{\tabcolsep}{8pt}
\begin{tabular}{cccc}
\toprule  & Rhino midden & Termite mound & Water \\
\midrule Thermal & $0.35\pm0.02$ & $0.30\pm0.03$ & $\mathbf{0.40\pm0.03}$ \\
 RGB & $0.25\pm0.02$ & $0.27\pm0.03$ & $0.24\pm0.03$ \\
 LiDAR & $\mathbf{0.40\pm0.02}$ & $\mathbf{0.43\pm0.04}$ & $0.36\pm0.04$\\
\bottomrule
\end{tabular}
\label{table:gating_weights}
\end{table}

\section{Conclusion}
\label{sec:conclusion}
We studied three methods for fusing thermal, RGB, and LiDAR imagery for mapping three biophysical landscape features: rhino middens, termite mounds, and water.
\textsc{Early fusion} concatenates multimodal tiles of the same resolution and passes them through a 5-channel ResNet. \textsc{Late fusion} passes tiles of different modalities through separate ResNet feature extractors before fusing them in a fully connected layer. \textsc{Mixture of Experts} uses a gating network to weight the predictions from each modality as a function of the input.
Overall, the results show that while the three methods have similar macro-averaged performance with \textsc{Late fusion} achieving an AUC of 0.698, their per-class performance varies strongly, with \textsc{Early fusion} achieving the best recall for middens and water and \textsc{Mixture of Experts} achieving the best recall for mounds.

\section*{Acknowledgments} 
We acknowledge South African National Parks for logistical and scientific support, and for permission to perform the study in Kruger National Park.
This work was supported in part by the Pioneer Centre for AI, DNRF grant number P1. L.G. was supported by the National Science Foundation Graduate Research Fellowship under Grant No. DGE2140743.

\bibliographystyle{splncs04}
\bibliography{references}

\begin{thebibliography}{10}
\providecommand{\url}[1]{\texttt{#1}}
\providecommand{\urlprefix}{URL }
\providecommand{\doi}[1]{https://doi.org/#1}

\bibitem{burke}
Burke, M., Driscoll, A., Lobell, D.B., Ermon, S.: \href{https://www.science.org/doi/10.1126/science.abe8628}{Using satellite imagery to understand and promote sustainable development}. Science  \textbf{371}(6535) (2021). \doi{10.1126/science.abe8628}

\bibitem{imagenet}
Deng, J., Dong, W., Socher, R., Li, L.J., Li, K., Fei-Fei, L.: \href{https://ieeexplore.ieee.org/document/5206848}{ImageNet: A large-scale hierarchical image database}. In: 2009 IEEE Conference on Computer Vision and Pattern Recognition. pp. 248--255 (2009). \doi{10.1109/CVPR.2009.5206848}

\bibitem{UN}
of~Economic, U.N.D., Affairs, S.: \href{https://sdgs.un.org/goals/goal15}{Goal 15: Protect, restore and promote sustainable use of terrestrial ecosystems, sustainably manage forests, combat desertification, and halt and reverse land degradation and halt biodiversity loss} (2023), accessed: 2023-05-24

\bibitem{eigen}
Eigen, D., Ranzato, M.A., Sutskever, I.: \href{https://arxiv.org/pdf/1312.4314}{{Learning Factored Representations in a Deep Mixture of Experts}}. ICLR  (2014)

\bibitem{gordon}
Gordon, L., Behari, N., Collier, S., Bondi-Kelly, E., Killian, J.A., Ressijac, C., Boucher, P., Davies, A., Tambe, M.: \href{https://www.ijcai.org/proceedings/2023/0663.pdf}{Find Rhinos without Finding Rhinos: Active Learning with Multimodal Imagery of South African Rhino Habitats}. In: Proceedings of the Thirty-Second International Joint Conference on Artificial Intelligence, {IJCAI-23}. pp. 5977--5985. International Joint Conferences on Artificial Intelligence Organization (8 2023). \doi{10.24963/ijcai.2023/663}, aI for Good

\bibitem{he}
He, K., Zhang, X., Ren, S., Sun, J.: \href{https://arxiv.org/pdf/1512.03385}{Deep Residual Learning for Image Recognition}. In: 2016 IEEE Conference on Computer Vision and Pattern Recognition (CVPR). pp. 770--778 (2016). \doi{10.1109/CVPR.2016.90}

\bibitem{kang}
Kang, B., Xie, S., Rohrbach, M., Yan, Z., Gordo, A., Feng, J., Kalantidis, Y.: \href{https://arxiv.org/pdf/1910.09217}{Decoupling Representation and Classifier for Long-Tailed Recognition}. In: Proceedings of the The Eighth International Conference on Learning Representations, {ICLR-20}. International Conference on Learning Representations (2020). \doi{https://doi.org/10.48550/arXiv.1910.09217}

\bibitem{loveridge}
Loveridge, J.P., Moe, S.R.: \href{https://www.cambridge.org/core/journals/journal-of-tropical-ecology/article/abs/termitaria-as-browsing-hotspots-for-african-megaherbivores-in-miombo-woodland/E2809392DD9ECB34C99A8FD0CA56EA4F}{{Termitaria as browsing hotspots for African megaherbivores in miombo woodland}}. Journal of Tropical Ecology  \textbf{20}(3),  337–343 (2004). \doi{10.1017/S0266467403001202}

\bibitem{owen1973behavioural}
Owen-Smith, R.N., Smith, R.N.O.: \href{http://www.rhinoresourcecenter.com/pdf_files/132/1320739004.pdf}{The Behavioural Ecology of the White Rhinoceros}. Ph.D. thesis, University of Wisconsin Madison (1973)

\bibitem{parrish}
Parrish, J.D., Braun, D.P., Unnasch, R.S.: \href{https://academic.oup.com/bioscience/article-pdf/53/9/851/26894780/53-9-851.pdf}{{Are We Conserving What We Say We Are? Measuring Ecological Integrity within Protected Areas}}. BioScience  \textbf{53}(9),  851--860 (09 2003). \doi{10.1641/0006-3568(2003)053[0851:AWCWWS]2.0.CO;2}

\bibitem{rolf}
Rolf, E., Klemmer, K., Robinson, C., Kerner, H.: \href{https://arxiv.org/pdf/2402.01444}{{Mission Critical -- Satellite Data is a Distinct Modality in Machine Learning}} (2024)

\bibitem{yeh}
Yeh, C., Perez, A., Driscoll, A., Azzari, G., Tang, Z., Lobell, D., Ermon, S., Burke, M.: \href{https://www.nature.com/articles/s41467-020-16185-w}{{Using publicly available satellite imagery and deep learning to understand economic well-being in Africa}}. Nature Communications  \textbf{11}(2583) (2020)

\end{thebibliography}
\end{document}